\PassOptionsToPackage{table,dvipsnames,xcdraw}{xcolor}
\documentclass[12pt]{style}
\usepackage[toc,page,header]{appendix}
\usepackage{wrapfig}

\usepackage{natbib}

\usepackage{CJKutf8}

\usepackage{xargs}

\usepackage{todonotes}

\usepackage{multirow}

\usepackage{cleveref}

\usepackage{amsmath}
\usepackage{amssymb}
\usepackage{mathtools}
\usepackage{amsthm}
\usepackage{dsfont}

\usepackage{svg}

\usepackage{mathrsfs}
\usepackage{adjustbox}
\usepackage{multicol}
\usepackage{tcolorbox}
\usepackage{changepage}
\usepackage{enumitem}
\usepackage{graphicx}
\usepackage{xcolor}
\usepackage{float}
\usepackage{threeparttable}
\usepackage{subcaption}
\usepackage{algorithm}
\usepackage{algpseudocode}
\usepackage{wrapfig}
\usepackage[table]{xcolor}
\usepackage{colortbl}
\usepackage{tabularx}
\usepackage{makecell}
\usepackage[dvipsnames]{xcolor}

\newcolumntype{g}{>{\columncolor{gray!10}}c}

\theoremstyle{plain}

\theoremstyle{definition}

\theoremstyle{remark}

\definecolor{catgray}{gray}{0.9}
\definecolor{skyblue}{rgb}{0.53,0.81,0.92}
\colorlet{skyblue!30}{skyblue!30!white}
\definecolor{customblue}{RGB}{70,130,180}

\newcommand{\cmark}{\textcolor{green!60!black}{\checkmark}}
\newcommand{\xmark}{\textcolor{red!70!black}{\ding{55}}}

\renewcommand{\emph}[1]{\textit{#1}}

\usepackage{minitoc}
\usepackage{url}

\PassOptionsToPackage{table,xcdraw}{xcolor}
\usepackage{titletoc}
\usepackage{placeins}
\usepackage{pifont}

\definecolor{RowBlue}{HTML}{E9F2FB}
\definecolor{RowRed}{HTML}{F9EAEA}
\definecolor{Top1}{HTML}{50DB4B}
\definecolor{Top2}{HTML}{A5FFA2}
\definecolor{Top3}{HTML}{D9FFD9}
\definecolor{Sub1}{HTML}{EAB8B8}
\definecolor{Sub2}{HTML}{E4E4E4}

\usepackage{listings}
\definecolor{codecomment}{RGB}{58,128,128}
\definecolor{codefunc}{RGB}{204,37,124}
\lstdefinestyle{fdcode}{
  language=Python,
  basicstyle=\ttfamily\footnotesize,
  columns=fullflexible,
  keepspaces=true,
  commentstyle=\color{codecomment},
  keywordstyle={},
  stringstyle={},
emph={G,S,Dec,phi,psi,FD,FD_ema,all_gather,flow_matching,renoise,sample_clip,backward,step,detach},
  emphstyle=\color{codefunc},
  showstringspaces=false,
  aboveskip=2pt,
  belowskip=0pt,
}

\newcommand{\FD}{\mathrm{FD}}
\newcommand{\sg}{\operatorname{sg}}
\newcommand{\runin}[1]{{\bfseries #1}}
\makeatletter
\def\ff@sans{sans}
\newcommand{\ff@sfeacute}{\leavevmode\setbox0\hbox{e}%
  \setbox2\hbox{\fontfamily{phv}\fontseries{b}\selectfont x}%
  \rlap{\hbox to\wd0{\hss\raisebox{\dimexpr\ht0-\ht2\relax}{\fontfamily{phv}\fontseries{b}\selectfont\'{}}\hss}}e}
\DeclareRobustCommand{\eacute}{\ifx\f@family\ff@sans\ff@sfeacute\else\'e\fi}
\makeatother
\newcommand{\Frechet}{\texorpdfstring{Fr\eacute chet}{Fréchet}}
\newcommand{\method}{\Frechet{} Forcing}
\newcommand{\ours}{MiniWorld-1B-FF}
\newcommand{\best}[1]{{\rmfamily\bfseries #1}}

\title{\method: Training Long-Horizon Streaming\\
World Models from Short Clips}

\author{Yian Zhao}
\author{Yu Yan}
\author{Hongcan Guo}
\author{Ruochong Zheng}
\author{Jian Zhang}
\author{Jie Chen}

\newcommand{\githubicon}{\raisebox{-0.15em}{\includegraphics[height=1.05em]{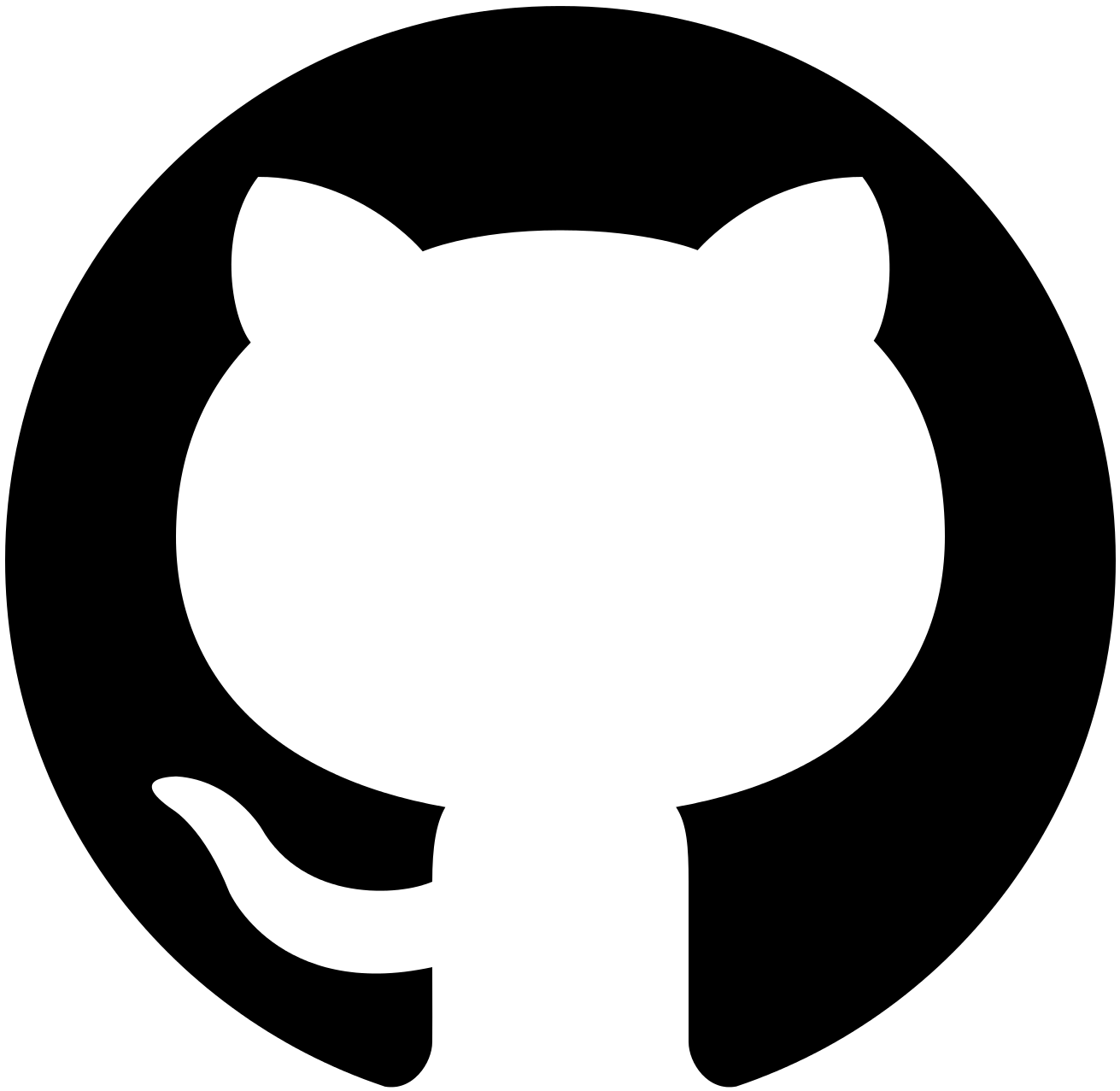}}}
\newcommand{\hficon}{\raisebox{-0.15em}{\includegraphics[height=1.05em]{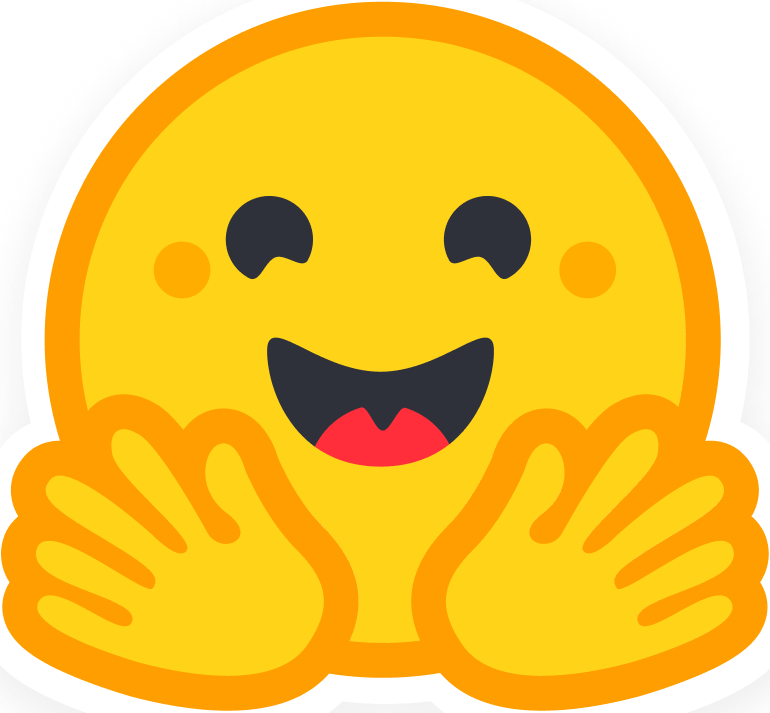}}}
\newcommand{\globeicon}{\raisebox{-0.15em}{\includegraphics[height=1.05em]{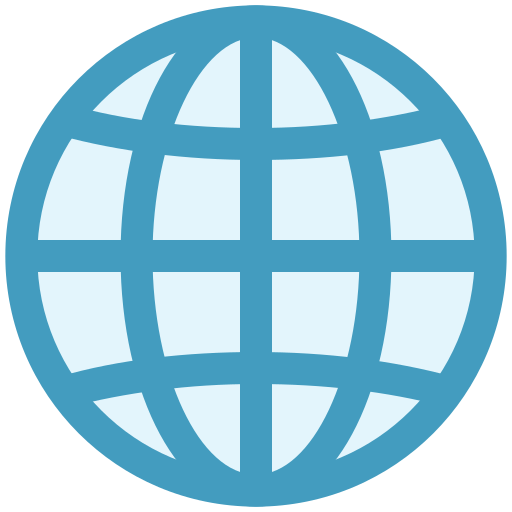}}}
\newcommand{\resourcelink}[1]{\href{#1}{\texttt{\detokenize{#1}}}}

\abstract{
Streaming video world models generate the future chunk by chunk, conditioned on
their own past and a stream of actions. When trained on short clips, they drift
once a rollout exceeds the training horizon, as prediction errors re-enter the
context and compound. A common remedy is to train the model on its own rollouts
via distribution matching distillation (DMD). For action-conditioned world
models, however, the teacher becomes a bottleneck: an external video model is
agnostic to the actions, whereas the model's own pretrained network drifts in
the same manner as the student. We observe that a long rollout need not match a
teacher, provided it remains a realistic continuation of its own scene that
follows the commanded actions. Realism and scene fidelity can be measured
directly against real-data statistics, which remain valid at any horizon, while
action following is preserved by the model's conditional generation loss.
Building on this insight, we propose \method{}, which fine-tunes a world model
together with its deployed multi-step sampler under three complementary
constraints. \emph{Generative anchoring} applies the flow-matching loss after a
self-generated history ending on real data, preserving action-conditioned
generation. \emph{Context fidelity}, a Fr\'echet distance (FD) on latent
statistics of the free rollout centred on each clip's own ground truth, keeps
every rollout faithful to its scene. \emph{Reality matching}, an FD to real
frames in a frozen perceptual space, keeps the generated frames realistic.
Because all three targets are derived from short real clips without comparing a
rollout against a long recorded future, training requires no teacher, no online
critic and no long videos. Fine-tuning MiniWorld-1B, pretrained on 253-frame
clips, for at most 36 GPU-hours on 381-frame clips yields rollouts that remain
stable beyond 4,000 frames and continue to follow their actions in both robot
manipulation (DROID) and camera-controlled exploration (Sekai). On DROID, PSNR
against a forward-backward episode replay stays above 21\,dB through frame
4,000, compared with 8.9\,dB for the base model. An analysis of the training
dynamics further reveals how distribution matching on free rollouts fails and
how each design choice prevents these failure modes.
}

\checkdata[\githubicon\hspace{0.45em}Code]{\resourcelink{https://github.com/zhao-yian/Frechet_Forcing}}
\checkdata[\hficon\hspace{0.45em}Models]{\resourcelink{https://huggingface.co/zhaoyian01/Frechet_Forcing}}
\checkdata[\globeicon\hspace{0.45em}Page \& Demo]{\resourcelink{https://zhao-yian.github.io/Frechet_Forcing}}

\begin{document}
\maketitle

%=============================================================================
\section{Introduction}
\begin{figure}[tbp]
\centering
\includegraphics[width=\linewidth]{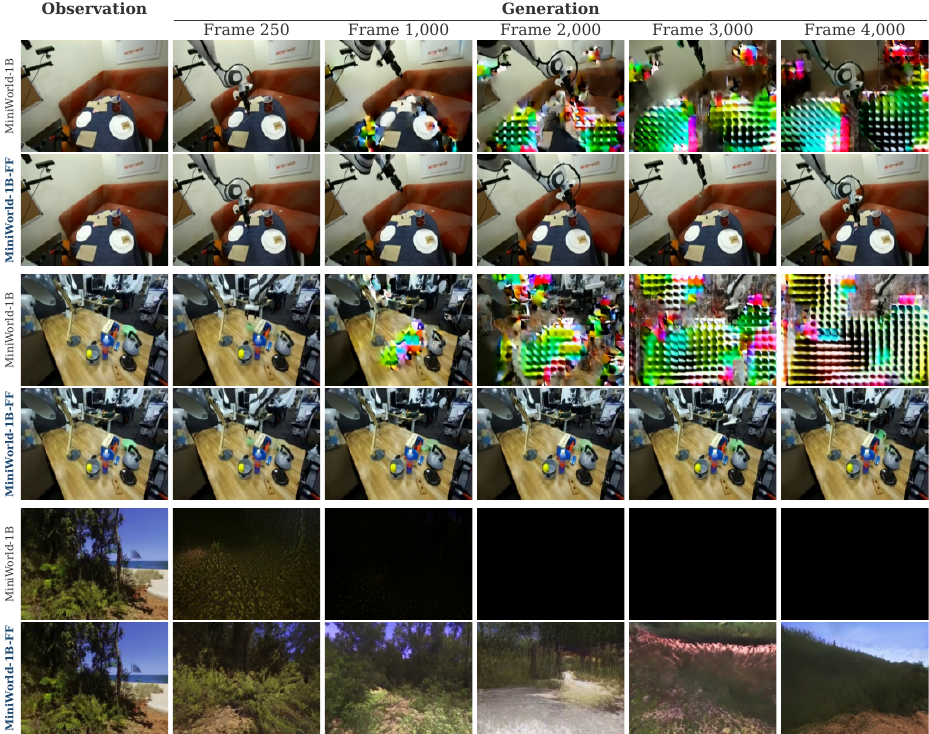}
\caption{4,000-frame chunk-by-chunk rollouts from a single real frame on
two DROID episodes (top four rows) and one Sekai scene
(bottom two rows), following the commanded robot or camera trajectory.
MiniWorld-1B, pretrained on 253-frame clips, cannot sustain longer
rollouts: on DROID, saturated lattice texture spreads from the moving arm
within 1,000 frames, and on Sekai the rollout darkens within the first 250
frames and then fades to black. \ours{}, fine-tuned with
\method{} on 381-frame clips, stays coherent to frame 4,000, over
$10\times$ the fine-tuning horizon.}
\label{fig:teaser}
\end{figure}

%=============================================================================
World models that generate video in a streaming fashion, with each chunk
conditioned on previously generated chunks and on an action, provide a natural
substrate for interactive simulation, game engines and embodied
agents~\cite{ha2018worldmodels,bruce2024genie,alonso2024diamond,valevski2025gamengen,chen2024diffusionforcing,yin2025causvid,huang2025selfforcing}.
However, because they are trained on ground-truth history and deployed
autoregressively on their own predictions~\cite{bengio2015scheduled}, small
errors re-enter the context and compound over time. Rollouts therefore remain
coherent for roughly the length of the training clips before drifting, fading
to black or dissolving into texture (Fig.~\ref{fig:teaser}). This long-horizon
drift remains the central obstacle to sustained streaming
generation~\cite{huang2025selfforcing,cui2025selfforcingpp,yang2025longlive}.

The prevailing strategy to close this train-test gap is to train the model on
its own rollouts. Because a self-generated rollout has no paired ground truth
beyond the first few seconds, supervision must be distributional, and recent
work has largely converged on distribution matching distillation
(DMD)~\cite{yin2024dmd,yin2024dmd2,yin2025causvid,huang2025selfforcing}, where the
difference between a frozen teacher's score and an online ``fake'' score of the
student on noised student samples provides the training gradient. State-of-the-art
systems reach minute-scale video by extending DMD to longer rollouts, rolling
windows and chunk-wise
scoring~\cite{cui2025selfforcingpp,liu2025rolling,yang2025longlive,gao2026rmd}.
Teacher-free alternatives instead retain a per-sample regression target by
conditioning the flow-matching or diffusion loss on a history corrupted by the
model's own errors~\cite{guo2025resampling,li2026svi,yang2026arforcing}, or by
reversing rollouts so that they end on or recover the ground-truth
start~\cite{po2026bagger,huang2026live}.

For action-conditioned world models, however, the teacher poses a fundamental
dilemma. An effective teacher must simultaneously model the student's domain
and action space while remaining more accurate than the student \emph{on the
states the student visits}. A large external video model violates the first
requirement, as it is agnostic to the action signal and pulls rollouts towards
a mismatched distribution. Conversely, the world model's own pretrained network
satisfies the first requirement but violates the second: it is the very model
whose rollouts drift, so on long-horizon states its score agrees with the
student and provides no corrective signal. In our experiments, neither choice
improved over the base model across a broad range of configurations. Per-sample
teacher-free objectives avoid the teacher altogether, yet they supervise only
states that can be paired with ground truth, where a degraded history is
followed by its recorded continuation or a reversed rollout recovers its
recorded start. Because the regression target always lies within a recorded
clip, the states a rollout visits thousands of frames later are never supervised
directly.

We take a different view of the supervision required for long rollouts. Rather
than matching a teacher or a paired recording, a rollout must remain a
realistic continuation of its own scene while following the commanded actions.
These properties call for complementary training signals. Visual realism and
scene fidelity are distributional properties of the rollout that can be
evaluated against real data at any horizon, because the distribution of real
frames is fixed, compact to summarise and, unlike a learned score, equally valid
one second or one minute into a rollout. Action following, by contrast, is a
property of conditional generation that the model's pretraining loss already
supervises. \method{} therefore fine-tunes free rollouts generated directly by
the deployed sampler to minimise Fr\'echet distances (FD) to real-data
statistics while retaining the conditional generation loss. Running moment
estimates~\cite{yang2026fdloss} render the FD differentiable at a batch size of
one per GPU, and truncated backpropagation through the final chunks keeps
rollout training computationally tractable. Specifically, our objective
combines three constraints, each targeting one essential property of a long
rollout. \emph{Generative anchoring} preserves action-conditioned generation by
applying the flow-matching (FM) loss after a self-generated history ending on
real data, exposing FM to the model's accumulated errors while keeping the
regression target exact. \emph{Context fidelity} keeps each rollout faithful to
its initial scene through an FD on latent statistics centred on the clip's own
ground truth. \emph{Reality matching} enforces frame-level realism through an FD
to real frames in a frozen self-supervised feature space, penalising the dark,
washed-out and tiled artifacts towards which long rollouts otherwise drift.

Because none of the three constraints compares a rollout against a long
recorded future, long-horizon training no longer requires long videos.
Generative anchoring regresses a short real window after a re-rendered history,
context fidelity compares the final chunks of a 317-frame rollout against the
corresponding span of its own clip, and reality matching compares generated
frames against a population of real frames. Every training sample is a
381-frame clip, and long-horizon drift is suppressed by supervising the states
that the model reaches and learns to correct within these rollouts. At
deployment, the rollout extends for as long as the action stream that drives it.
We validate \method{} on MiniWorld-1B~\cite{zhao2026miniworld}, a block-causal
world model pretrained on 253-frame clips, across two distinct domains. After
fine-tuning on 381-frame clips, its rollouts remain stable beyond 4,000 frames,
over $10\times$ the pretraining horizon, in both robot manipulation on DROID and
camera-controlled exploration on Sekai (Fig.~\ref{fig:teaser}).

Naive distribution matching on free rollouts fails in instructive ways. We
analyse its training dynamics with three diagnostic probes that track whether
outputs remain dependent on their context, whether gradients align with the
required statistical correction, and whether an external perceptual judge agrees
with the training objective (\S\ref{sec:analysis}, Fig.~\ref{fig:dynamics}).
This analysis reveals three distinct failure modes: matching a marginal FD
pulls every rollout towards a dataset-average appearance (\emph{context
decoupling}), pooling statistics over the entire rollout forces the
gradient-carrying tail to compensate for the detached prefix
(\emph{gradient-statistic misalignment}), and matching a low-order feature
space admits degenerate frames that do not resemble real images
(\emph{feature-space exploitation}). The critical design choices are therefore
\emph{where} a rollout is scored and \emph{in which feature space}: context
fidelity scores clip-centred statistics solely on the gradient-carrying chunks,
while reality matching provides a semantic feature space, making the two
distributional constraints mutually complementary. Our analysis also uncovers a
failure mode specific to training on self-generated history: when a re-rendered
history does not meet the real target smoothly, the model learns to redraw the
scene at every chunk boundary. We quantify this periodic redraw with a dedicated
chunk-boundary flicker metric and eliminate it by ending the history on a single
real chunk.

In summary, our contributions are threefold:
\begin{itemize}
\item We propose \method{}, to our knowledge the first method that trains a
streaming video world model on its own rollouts using Fr\'echet-distance
objectives against real data. It requires no teacher, no online critic and no
long videos, and fine-tunes the deployed multi-step sampler without
modification.
\item We analyse distribution matching on free rollouts to identify three
distributional failure modes (context decoupling, gradient-statistic
misalignment and feature-space exploitation) alongside chunk-boundary redraw
induced by self-generated history, and derive the exact design choice that
resolves each.
\item Using a single hyperparameter recipe and at most 36 GPU-hours of
fine-tuning on 381-frame clips, \method{} extends the stable horizon of a 1B
world model pretrained on 253-frame clips beyond 4,000 frames in two domains,
reducing last-bin FVD by $25\times$ on DROID and $2.7\times$ on Sekai while
preserving action following.
\end{itemize}

%=============================================================================
\section{Related Work}
%=============================================================================
\paragraph{Autoregressive and streaming video generation.} Large-scale video
diffusion transformers~\cite{peebles2023dit,yang2025cogvideox,kong2024hunyuanvideo,wan2025}
synthesise full clips with bidirectional attention, whereas interactive world
models and neural game
engines~\cite{ha2018worldmodels,hafner2025dreamerv3,bruce2024genie,alonso2024diamond,valevski2025gamengen}
must generate observations causally under online control. Diffusion
Forcing~\cite{chen2024diffusionforcing} assigns independent noise levels to
chunks so that a model can denoise the future while conditioning on a cleaner
past, forming the foundation of most streaming world models.
CausVid~\cite{yin2025causvid} and Self Forcing~\cite{huang2025selfforcing} train
on self-generated history with DMD~\cite{yin2024dmd,yin2024dmd2}, distilling a
bidirectional teacher into a few-step causal student.
MiniWorld~\cite{zhao2026miniworld} is instead a causal world model trained from
scratch, using a block-causal DiT trained with diffusion forcing under a
non-decreasing chunk noise schedule and deployed with a rolling KV cache.
Action-conditioned video models also serve as simulators for navigation and
robot manipulation~\cite{yang2024unisim,bar2025nwm,agarwal2025cosmos}. Because
they are trained on short recorded episodes, a rollout that extends far beyond
an episode has neither training data nor a recorded ground-truth reference.

\paragraph{Long-horizon generation.} Training-free methods act on the context
through attention sinks~\cite{xiao2024sinks}, rolling windows and diagonal
schedules~\cite{ruhe2024rolling,kim2024fifo}, history
compression~\cite{zhang2025framepack}, history
guidance~\cite{song2025history} or spectral anchoring~\cite{freqforcing}. Our
sampler already incorporates an attention sink, a rolling cache and a staircase
schedule, yet still drifts without fine-tuning. Training-based methods
predominantly build on DMD. Self-Forcing++~\cite{cui2025selfforcingpp}, Rolling
Forcing~\cite{liu2025rolling} and LongLive~\cite{yang2025longlive} train on
rollouts longer than the teacher's clips by scoring windowed segments.
OPSD-V~\cite{liu2026opsdv} supplies the teacher with a cleaner cache,
CMD~\cite{bandyopadhyay2026cmd} employs a causal teacher conditioned on the
student's context, and RMD~\cite{gao2026rmd} scores each chunk independently so
that the teacher does not preserve artifacts from the generated history. All of
these methods rely on a pretrained teacher and an online critic to train a
few-step generator. Teacher-free methods instead retain the diffusion or
flow-matching loss while modifying its conditioning. The first family conditions
the loss on a history corrupted by the model's own errors: Resampling
Forcing~\cite{guo2025resampling} re-noises the ground-truth history and denoises
it with the model itself, Stable Video Infinity~\cite{li2026svi} injects banked
model errors into clean inputs, and AR Forcing~\cite{yang2026arforcing} feeds
the model's predictions back into the context of a navigation world model. The
second family exploits time reversal: BAgger~\cite{po2026bagger} reverses the
model's drifted rollouts into corrective trajectories ending on real frames, and
LIVE~\cite{huang2026live} reverses a rollout under reversed actions to apply the
loss against the recovered ground-truth start. In both families, the regression
target remains a recorded frame, restricting supervision to states that can be
paired with ground truth.

\paragraph{Distributional objectives.} FID~\cite{heusel2017fid} and
FVD~\cite{unterthiner2018fvd} are standard evaluation metrics for generative
models. Optimising them directly was long considered impractical because
estimating a high-dimensional covariance requires far more samples than a single
training batch provides. Yang et al.~\cite{yang2026fdloss} decouple the
population size from the batch size using running moment estimates and show that
the resulting loss effectively trains image generators. We bring this principle
to streaming video world models, where the samples are rollout chunks, gradients
are truncated in time, and the primary failure modes are temporal.

%=============================================================================
\section{\method}
\label{sec:method}
%=============================================================================
\subsection{Setting}
Our experiments build on the open-source MiniWorld codebase and its released
checkpoints~\cite{zhao2026miniworld}. We consider a block-causal video
DiT~\cite{peebles2023dit} over VAE latents that generates latent chunks
conditioned on a KV cache of past chunks and per-frame actions, such as camera
poses or robot commands. The model
is pretrained with diffusion forcing and deployed with a streaming sampler that
maintains several chunks in flight at staggered noise levels. Our only
requirement is that this sampler is differentiable, and neither the model
architecture nor the sampler is modified. The training data consist of short
real clips paired with actions, each slightly longer than a single training
rollout, whereas deployment rollouts are unbounded in length.

\subsection{Three constraints}
A long rollout remains effective only if it satisfies three requirements that no
single objective can enforce alone. First, the model must remain a faithful
\emph{conditional generator}, where every chunk continues its history and
follows its commanded action. Second, each rollout must remain faithful to
\emph{its own scene}: a trajectory starting in a dusk forest should still depict
that forest a thousand frames later rather than collapsing to a dataset-average
appearance. Third, the generated frames must remain \emph{realistic}, free from
darkening, washing out or periodic tiling. \method{} addresses each requirement
with a dedicated constraint,
\begin{equation}
\mathcal L = \underbrace{\mathcal L_{\mathrm{anc}}}_{\text{generative anchoring}}
+ \underbrace{\mathcal L_{\mathrm{ctx}}}_{\text{context fidelity}}
+ \underbrace{\mathcal L_{\mathrm{real}}}_{\text{reality matching}},
\label{eq:objective}
\end{equation}
combined with unit weights throughout. Generative anchoring is a regression loss
on real clips applied over a short window of chunks, whereas context fidelity
and reality matching are distributional losses applied directly to free rollouts
of the deployed sampler, where long-horizon drift actually emerges.

\subsection{Generative anchoring}
\label{sec:anchoring}
Fine-tuning on rollout statistics alone allows the model to drift away from its
core role as an action-conditioned generator that continues a given history.
Generative anchoring retains the pretraining flow-matching (FM)
objective~\cite{lipman2023flow,liu2023rectified}, but conditions it on a history
that carries the model's own prediction errors, following the self-resampling
scheme of Resampling Forcing~\cite{guo2025resampling} with one essential
modification regarding how the history terminates. From a real clip $x$, we draw a history length
$c\sim U[1,N]$, noise the first $c$ chunks to $t_a=0.7$ and denoise them with
the deployed staircase sampler without gradients. Because each chunk retains
roughly 30\% of its clean signal, its content remains anchored to the clip, yet
being re-rendered autoregressively on preceding re-renderings imbues the history
with the model's own rollout errors. The FM loss is then applied to the
subsequent $W=4$ real chunks:
\begin{equation}
\mathcal L_{\mathrm{anc}} = \mathbb E_{t,\epsilon}\,
\big\|v_\theta(x^{t}_{c:c+W}\mid h_c, a) - (\epsilon - x_{c:c+W})\big\|^2,
\qquad
h_c = \big[\,\sg\,S(G, x_{<c-1}^{t_a}, a),\; x_{c-1}\big],
\end{equation}
where $x^t=(1-t)\,x+t\,\epsilon$ with per-chunk noise levels drawn as in
pretraining, $\sg$ denotes the stop-gradient operator, and $S(G,\cdot)$ runs the
deployed sampler with model $G$ starting from noise level $t_a$.

\paragraph{The history must end on real data.} Crucially, the final chunk of
$h_c$ is the clean real chunk $x_{c-1}$ rather than its re-rendered counterpart
$\tilde x_{c-1}$. At noise level $t_a=0.7$, the model alters mid-scale spatial
structures such as the outline of a tree canopy or a cloud bank, causing
$\tilde x_{c-1}$ to deviate from $x_{c-1}$ at the scale of image blocks. Because
the FM target is the true continuation of $x_{c-1}$, which cannot be inferred
consistently from $\tilde x_{c-1}$, the loss-minimising strategy under a fully
re-rendered history is \emph{not} to continue the preceding frames smoothly, but
to redraw a fresh version of the scene. A model trained in this manner learns to
redraw the scene at every chunk boundary, producing severe block-scale flicker at
the chunk period (\S\ref{sec:seam}). Because temporal continuity across the seam
is a local relation between the target window and the end of its history,
restoring only the final chunk $x_{c-1}$ resolves this contradiction while the
preceding $c-1$ chunks still expose the model to its own errors. Ending the
history on real data also lowers the median FM loss from 0.26 to 0.15 over the
first 150 training steps, confirming that a large fraction of the loss was
previously dominated by this irreducible boundary mismatch.

\subsection{Distribution matching on free rollouts}
Context fidelity and reality matching share a common rollout mechanism. At each
training step, the model rolls out $N$ chunks autoregressively from a single
real context latent using the deployed sampler. Only the final $K$ sampler steps
are retained in the autograd graph, and completed chunks enter the KV cache
detached. Gradients therefore flow only into the last $K$ chunks, reaching the
newest chunk through its full denoising trajectory and earlier chunks through
progressively shorter suffixes, without backpropagating through the history. The
detached prefix serves to bring the model to the states it visits during
deployment rather than acting as a direct optimisation target. Each
distributional term embeds the gradient-carrying chunks with a frozen feature
extractor, gathers the resulting features across GPUs and matches their first
and second moments to those of real data.

Let $(\mu_r,\Sigma_r)$ denote the reference mean and covariance of a feature map
$\phi$ on real data, and let $\{f_i\}_{i=1}^n$ denote the features extracted
from the gradient-carrying rollout chunks across all GPUs at the current step,
with batch moments $\mu_{\mathrm{b}}=\tfrac1n\sum_{i=1}^n f_i$ and
$M_{\mathrm{b}}=\tfrac1n\sum_{i=1}^n f_i f_i^{\!\top}$. Because a single step
yields at most a few hundred feature vectors, which is insufficient for a
full-rank covariance estimate in hundreds of dimensions, we follow Yang et
al.~\cite{yang2026fdloss} and decouple the effective population size from the
batch size. Specifically, we maintain exponential moving averages (EMA) of the
first and second moments $(\bar\mu,\bar M)$ with decay rate $\beta$,
differentiate solely through the current batch, and recover the covariance
$\Sigma$ from the updated moments:
\begin{align}
\mu &= \beta\,\sg(\bar\mu) + (1-\beta)\,\mu_{\mathrm{b}}, &
M &= \beta\,\sg(\bar M) + (1-\beta)\,M_{\mathrm{b}}, \\
\Sigma &= M - \mu\mu^{\!\top}, &
\FD(\mu,\Sigma) &= \|\mu-\mu_r\|_2^2 + \mathrm{tr}\big(\Sigma+\Sigma_r-2(\Sigma_r^{1/2}\Sigma\Sigma_r^{1/2})^{1/2}\big).
\end{align}
After each step, the running buffers are updated as
$(\bar\mu,\bar M)\leftarrow(\sg(\mu),\sg(M))$, providing an effective sample
horizon of $1/(1-\beta)$ steps without storing past features. Each
distributional constraint is then optimised via the self-normalised objective
$\mathcal L_\phi = \FD/\sg(\FD)$ (\texttt{fd / fd.detach()} in
Algorithm~\ref{alg:ff})~\cite{yang2026fdloss}. Detaching the denominator fixes
the forward value to $1$ and yields the backward gradient
$\nabla\FD/\sg(\FD) = \nabla\log\FD$, which preserves the gradient direction of
$\FD$ while normalising its magnitude by the current distance. This
self-normalisation makes the loss invariant to the scale and dimensionality of
$\phi$, prevents the distributional gradients from vanishing relative to
$\mathcal L_{\mathrm{anc}}$ as $\FD$ decreases, and allows the three terms in
Eq.~\eqref{eq:objective} to be combined with unit weights.

\subsection{Reality matching}
\paragraph{Enforcing frame-level realism.} To compute $\mathcal L_{\mathrm{real}}$,
the last $D\le K$ chunks are decoded to pixels, temporally sub-sampled and
embedded using the CLS token of a frozen DINOv2
ViT-B/14~\cite{oquab2023dinov2}, where $\mathcal L_{\mathrm{real}}$ is the
self-normalised FD between these embeddings and those of real frames. Because a
self-supervised perceptual representation captures semantic structure rather than
low-level pixel statistics, dark, washed-out or periodically tiled frames lie far
from the real distribution, ruling out the degenerate solutions that satisfy
low-order moments (\S\ref{sec:feature}). Since generated latents pass through
the VAE decoder before reaching DINOv2, we compute the reference moments on VAE
\emph{reconstructions} of real latents, decoded with the exact same window span
and temporal stride as during training, rather than on raw video. Using raw video
as the reference would include the VAE's reconstruction gap, which the latent
generator cannot close and would otherwise attempt to compensate for by
hallucinating high-frequency artifacts. Furthermore, because the VAE decoder is
temporal and causal, the beginning of a decoded window behaves differently from
subsequent frames: the first latent decodes to a single frame rather than four,
and the second latent reflects cache warm-up. We therefore decode one extra chunk
immediately preceding the scored window and discard its frames in both rollouts
and reference precomputation, ensuring that all scored frames share identical
decoder statistics.

\subsection{Context fidelity}
\paragraph{Preserving scene identity.} Because reality matching operates on the
marginal distribution of frames across clips, it is satisfied whenever the
pooled frames look realistic as a population and cannot verify whether an
individual rollout remains faithful to its initial scene. Context fidelity
enforces scene consistency by conditioning the statistical comparison on each
clip. Every latent in the gradient-carrying chunks is summarised by the spatial
mean and standard deviation of each channel, where the standard deviation makes
the representation sensitive to both spatial smoothing and high-frequency
noise. Summaries of two consecutive latents are concatenated into a single
feature vector $\psi$, so that their cross-latent covariance captures temporal
evolution and penalises frozen, flickering or discontinuous transitions that
single-latent statistics would overlook. We then \emph{centre} the rollout
features by subtracting the mean feature of the clip's own ground truth over the
corresponding temporal span,
\begin{equation}
h = \psi(z_{N-K:N}) - \overline{\psi(x_{N-K:N})},
\end{equation}
and define $\mathcal L_{\mathrm{ctx}}$ as the self-normalised FD between $h$ and
real video segments of the same span, each centred on its own segment mean.
Centring converts a marginal population metric into a clip-conditional one. A
rollout that remains faithful to its scene exhibits within-clip deviations
distributed like those of natural video around its mean, whereas a drifting
rollout produces systematic residuals: consistent biases across clips, such as
global darkening, shift the mean term, while clip-specific drifts inflate the
covariance beyond the natural within-segment variation of real video. Crucially,
neither error can be reduced by pulling rollouts towards a dataset-average
appearance, because every residual is measured relative to the clip's own ground
truth. Operating in latent space is a practical convenience rather than a strict
requirement, as applying the same clip-centred FD to low-order pixel statistics
of decoded frames is also effective (\S\ref{sec:ablation}). Algorithm~\ref{alg:ff}
summarises a complete training step.

\begin{algorithm}[tbp]
\caption{One training step of \method{}.}
\label{alg:ff}
\begin{lstlisting}[style=fdcode]
# G: streaming world model;  S: deployed streaming sampler
# Dec: frozen VAE decoder;  phi: frozen DINOv2;  psi: latent moments
# ref_ctx, ref_real: real-data moments (mu_r, sig_r)
# ema_ctx, ema_real: running 1st and 2nd moments (mu_bar, M_bar);  beta: EMA decay

x, a = sample_clip()                    # real latents and their actions

# Generative anchoring: FM after a self-denoised history
c = randint(1, N)
hist = S(G, renoise(x[:c], t=t_a), a).detach()
hist[-1] = x[c-1]                       # the history ends on real data
L_anc = flow_matching(G, x[c:c+W], hist, a)

# Free rollout from one real latent; only the last K chunks carry gradients
z = S(G, x[0], a)[-K:]
x_z = x[N-K:N]                          # ground truth over the same span

# Context fidelity: latent moments centred on the clip's own ground truth
L_ctx = FD_ema(psi(z) - psi(x_z).mean(0), ref_ctx, ema_ctx)

# Reality matching: perceptual moments of decoded frames
L_real = FD_ema(phi(Dec(z)), ref_real, ema_real)

(L_anc + L_ctx + L_real).backward()
optimizer.step()

# Normalised FD with EMA of 1st and 2nd moments
def FD_ema(f, ref, ema):
    f = all_gather(f)                   # gather features across GPUs
    mu_b, M_b = f.mean(0), f.T @ f / len(f)
    mu = beta * ema.mu.detach() + (1 - beta) * mu_b
    M  = beta * ema.M.detach()  + (1 - beta) * M_b
    sig = M - mu @ mu.T                 # covariance from 1st and 2nd moments
    ema.mu, ema.M = mu.detach(), M.detach()
    fd = FD((mu, sig), ref)
    return fd / fd.detach()
\end{lstlisting}
\end{algorithm}

\subsection{Relation to prior training signals}
\label{sec:relation}
Table~\ref{tab:dmd} compares \method{} with prior approaches that train on
self-generated rollouts or histories. DMD-based
methods~\cite{huang2025selfforcing,cui2025selfforcingpp,yang2025longlive,gao2026rmd}
define supervision through the score of a pretrained teacher. Consequently, the
training signal is valid only on states where the teacher itself is reliable,
requires an online critic network, and is tailored to few-step distillation. RMD
modifies \emph{where} the teacher is evaluated by scoring individual chunks, but
retains the teacher's score as the target. Teacher-free
methods~\cite{guo2025resampling,li2026svi,yang2026arforcing,po2026bagger,huang2026live}
replace the teacher with per-sample regression to ground truth, which restricts
supervision to states that can be paired with recorded frames. Specifically,
self-history methods regress the recorded continuation of a degraded history
while detaching the history itself, meaning that the rollout is exposed to the
loss but receives no direct gradient. Reversal-based methods require
time-reversible environment dynamics and apply the regression loss at the
recovered initial state rather than along the forward rollout. In contrast,
\method{} defines its targets via precomputed statistics of real data and
applies them directly to free rollouts of the deployed sampler, with gradients
propagating into the generated chunks. Because no frame-level pairing is
required, the distributional targets remain valid at arbitrary horizons, and all
auxiliary feature networks remain frozen in inference mode. Within our
framework, generative anchoring belongs to the self-history family, while
context fidelity and reality matching provide the direct rollout supervision
that per-sample regression lacks.

Two recent findings in the literature align closely with our formulation. First,
LIVE observes that sequence-level distribution matching permits unbounded
drift~\cite{huang2026live}. Our analysis traces this phenomenon to
\emph{marginal} distribution matching, which allows rollouts to collapse towards
a dataset-average appearance (\S\ref{sec:marginal}), and shows that context
fidelity prevents it by centring the statistics on each clip. Second, RMD argues
that rollout chunks should be scored independently of their degraded
context~\cite{gao2026rmd}. Reality matching achieves an analogous context-free
signal by scoring individual frames against a real frame population without
requiring a teacher.

\begin{table}[tbp]
\centering\small
\caption{Comparison of training signals for self-generated rollouts or histories.
Categories include DMD (Self Forcing~\cite{huang2025selfforcing}), long-horizon
DMD (Self-Forcing++~\cite{cui2025selfforcingpp}, Rolling
Forcing~\cite{liu2025rolling} and LongLive~\cite{yang2025longlive}),
RMD~\cite{gao2026rmd}, self-history (Resampling
Forcing~\cite{guo2025resampling}, SVI~\cite{li2026svi} and AR
Forcing~\cite{yang2026arforcing}), and reversal (BAgger~\cite{po2026bagger} and
LIVE~\cite{huang2026live}). GT denotes ground truth.}
\label{tab:dmd}
\footnotesize
\setlength{\tabcolsep}{3.65pt}
\begin{tabular*}{\linewidth}{l@{\hspace{2\tabcolsep}\extracolsep{\fill}}cccccc}
\toprule
 & DMD & long-horizon DMD & RMD & self-history & reversal & Ours \\
\midrule
\multirow{2}{*}{Target} & teacher & teacher score & per-chunk & GT & GT & real-data \\
 & score & on windows & teacher score & continuation & start & moments \\
\midrule
Teacher-free & \xmark & \xmark & \xmark & \cmark & \cmark & \cmark \\
Critic-free & \xmark & \xmark & \xmark & \cmark & \cmark & \cmark \\
Gradient into the rollout & \cmark & \cmark & \cmark & \xmark & \xmark & \cmark \\
Original sampler kept & \xmark & \xmark & \xmark & \cmark & \cmark & \cmark \\
\midrule
\multirow{2}{*}{Extra requirement} & \multirow{2}{*}{--} & \multirow{2}{*}{--} & chunk-adapted & \multirow{2}{*}{--} & reversible & reference \\
 & & & teacher & & dynamics & statistics \\
\bottomrule
\end{tabular*}
\end{table}

%=============================================================================
\section{Experiments}
\label{sec:exp}
%=============================================================================
We fine-tune MiniWorld-1B with \method{} using a single recipe across two
domains, robot manipulation on DROID and camera-controlled exploration on Sekai,
and evaluate both under a unified protocol up to frame 4,001. After describing
the experimental setup (\S\ref{sec:setup}), we evaluate whether long rollouts
remain realistic and stable (\S\ref{sec:horizon}) and whether they maintain
action controllability over the horizon (\S\ref{sec:control}).

\subsection{Setup}
\label{sec:setup}
\paragraph{Model.} MiniWorld-1B~\cite{zhao2026miniworld} is a block-causal DiT
over Wan2.2 VAE latents~\cite{wan2025} (48 channels, $15\times20$ spatial
latents for $240\times320$ frames, $4\times$ temporal compression) operating on
chunks of two latents. We start from the released domain-specific
checkpoints~\cite{zhao2026miniworld}, pretrained with diffusion forcing on clips
of up to 64 latents (253 frames). At inference, the model uses a staircase
sampler with 8 denoising steps per chunk, 8 chunks in flight, a rolling KV cache
of 24 chunks and a one-frame attention sink. Because the attended window never
exceeds the pretrained positional range, long-horizon degradation stems from
error accumulation in the context rather than positional extrapolation. Since
frame rate affects only playback speed, we report all horizons in frames.

\paragraph{Data.} \emph{DROID}~\cite{khazatsky2024droid} consists of robot
manipulation videos from the primary exterior camera at 15\,fps, conditioned on
the commanded absolute end-effector pose in the robot base frame and the gripper
state, normalised by dataset quantiles. Episodes are short, with a median length
of 209 frames. We fine-tune on the 8,958 successful episodes that are long
enough to form a training clip, and reserve 2,774 episodes for evaluation.
\emph{Sekai}~\cite{li2025sekai} consists of game-navigation videos paired with
camera poses, injected as per-pixel Pl\"ucker ray
encodings~\cite{he2025cameractrl}, recorded at 30\,fps and used without
temporal sub-sampling. We partition the dataset by source video
into 1,462 training and 156 held-out clips of 1,800 frames, computing all
reference statistics exclusively on the training split.

\paragraph{Training.} We fine-tune all model parameters using AdamW
($\beta=(0.9,0.95)$), a learning rate of $10^{-6}$ with a 100-step linear
warm-up followed by a constant schedule, and gradient norm clipping at 1.0, with
one 96-latent (381-frame) clip per GPU across 8 GPUs. These 381-frame clips
constitute the only ground-truth data used during fine-tuning and the longest
videos the model ever observes. Free rollouts span $N=40$ chunks (317 frames)
with gradients propagated through the final $K=8$ chunks. For reality matching,
the last $D=4$ chunks are decoded together with the immediately preceding chunk,
whose frames are discarded, and every second remaining frame is embedded at
$224\times308$ resolution. The latent reference statistics are computed on
16-latent real segments matching the span of the $K$ gradient-carrying chunks,
and the DINOv2 reference statistics are computed on VAE-reconstructed training
frames (52k latent windows from 4,000 episodes and 108k frames from 800 episodes
on DROID, and 26k latent windows from 2,000 segments and 54k frames from 400
clips on Sekai). Each FD estimator maintains running moments with $\beta=0.99$,
initialised from 16 gradient-free rollouts. Because each GPU rank differentiates
only through its local features before gradients are averaged across ranks, the
implemented loss scales by the world size $W$ as
$W\cdot\FD/(\sg(\FD)+0.01)$. For generative anchoring, the 4-chunk FM window
follows a history of $c\sim U[1,40]$ chunks obtained by noising the real clip to
$t_a=0.7$ and denoising it along the last four rungs of the deployed staircase,
with the final history chunk replaced by the clean ground-truth chunk. All
hyperparameters are identical across both domains. Fine-tuning requires only
225 steps, taking 4.2 hours on DROID and 4.5 hours on Sekai on 8 GPUs (34 and
36 GPU-hours at roughly 68\,s and 72\,s per step) with no teacher network in
memory.

\paragraph{Long-horizon evaluation protocol.} In each domain, we evaluate 32
held-out clips evenly spaced across the test set, rolling each out once from a
single real context latent for at least 1,001 latents (4,001 frames) using
identical action sequences and random seeds across all compared models. Because
no recording in either dataset reaches 4,000 frames, we construct long-horizon
action trajectories and ground-truth (GT) references as follows. \emph{DROID.}
Actions represent absolute end-effector poses under a fixed camera. Because a
pick-and-place trajectory played in reverse remains a valid manipulation
sequence, alternating an episode forwards and backwards yields a physically
plausible reference video of arbitrary length, whose frames in that order serve
as GT. We drive the model along the same forward-backward trajectory: forward
steps use the recorded commands directly, while a backward step targeting frame
$j$ uses the command recorded four frames prior to $j$ to account for the
empirical actuation lag. From the 481 held-out episodes of 240 to 600 frames
involving reversible object displacements (excluding actions such as pouring,
wiping, folding or opening), we roll out 32 evenly spaced episodes from their
initial frame for 1,024 latents (4,093 frames). This replay GT provides an
approximate reference, as a manipulated object need not return to its exact
initial pose after release. \emph{Sekai.} Each clip is rolled out from a single
context latent along its recorded 1,789-frame camera trajectory, and continued
beyond frame 1,789 by concatenating the relative camera motions of subsequent
held-out clips up to 1,001 latents, with GT available for the first 1,789
frames. A second protocol rolls out the same 32 clips for 1,789 frames along
their recorded trajectories with independent noise, providing full-length GT for
the action-following evaluation, ablations and training-dynamics analysis
(\S\ref{sec:analysis}).

\paragraph{Metrics.} All metrics are evaluated across six shared horizon bins by
pooling frames across the 32 rollouts: frames 1--253 (the pretraining horizon),
253--509, 509--1,021, 1,021--1,789, 1,789--2,997 and 2,997--4,001. None of the
evaluation backbones is used during training. CLIP-FD~\cite{radford2021clip}
embeds every eighth frame with CLIP ViT-B/16 and measures the FD against real
reference frames (all reconstructed frames of the 32 GT episodes on DROID, and
54k reconstructed training frames on Sekai). Because finite-sample FD depends on
sample size, we report each bin alongside the CLIP-FD of the GT frames at the
exact same sample count wherever GT is available. FVD is computed with
I3D~\cite{unterthiner2018fvd} over non-overlapping 16-frame windows against the
corresponding GT windows. Brightness is reported as the ratio of mean luma to
that of GT. Chunk-boundary flicker measures the mean absolute change in
$16\times16$-pooled luma across chunk boundaries relative to the change between
adjacent frames within a latent, averaged geometrically over rollouts
(\S\ref{sec:seam}), where GT scores 0.98--1.13. Action following is evaluated on
DROID via frame-wise PSNR against the replay GT, capturing preservation of the
scene, robot arm and manipulated objects, and on Sekai via camera trajectories
estimated by VGGT~\cite{wang2025vggt}. Specifically, on Sekai we measure both
the Pearson correlation $r$ between estimated and commanded yaw over 48-frame
windows and, on counterfactually forked left and right turn branches, the
percentage of branches that turn in the commanded direction and the fraction of
the commanded angle they achieve. All real references use VAE reconstructions so
that VAE compression error does not confound model comparisons. Finally, because
population metrics can mask rare localised failures, we additionally generate
4,001-frame rollouts on 96 evenly spaced held-out DROID episodes with \ours{}
(and eight with MiniWorld-1B) and manually inspect each video for visual
artifacts.

\paragraph{Baselines.} We compare \ours{} (MiniWorld-1B fine-tuned with
\method{}) against the pretrained MiniWorld-1B model and an FM-only baseline
that uses the exact same training clips, self-denoised history and schedule,
thereby isolating the contribution of the distributional constraints. On DROID,
FM-only uses the full generative anchoring formulation for the same 225 steps,
differing from \method{} solely by omitting $\mathcal L_{\mathrm{ctx}}$ and
$\mathcal L_{\mathrm{real}}$. On Sekai, FM-only uses generative anchoring with
the history re-rendered to the end (\S\ref{sec:seam}). We omit DMD baselines
because none of the DMD configurations we evaluated improved over the base
model.

\subsection{Realism and stability over 4,000 frames}
\label{sec:horizon}
\begin{figure}[tbp]
\centering
\includegraphics[width=\linewidth]{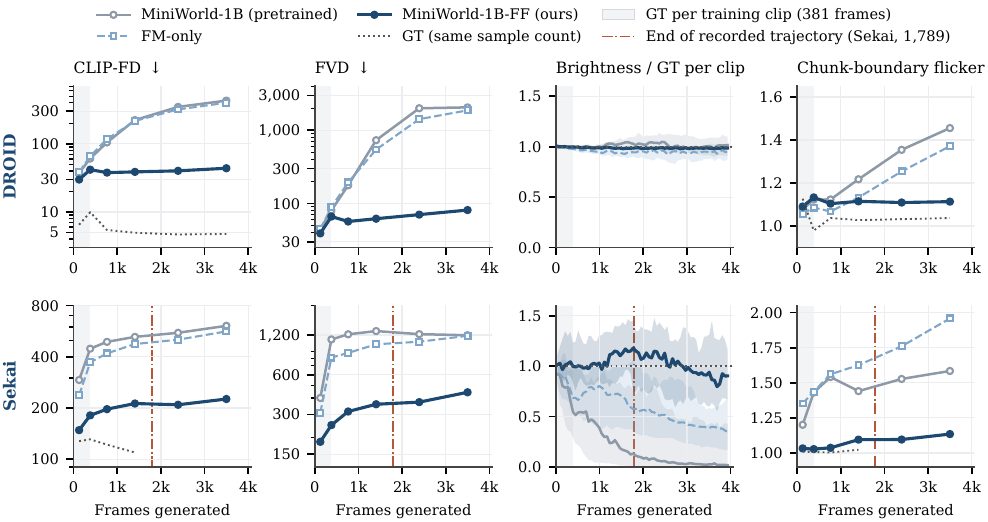}
\caption{Long rollouts on 32 held-out DROID episodes (top) and Sekai clips
(bottom), in the horizon bins of Table~\ref{tab:main}. CLIP-FD, FVD and
flicker are pooled over rollouts per bin, and dotted lines show GT at the same
sample count for CLIP-FD where GT exists. Brightness is computed per rollout
relative to the mean of its own GT in 60-frame windows, showing the median and
interquartile range over rollouts. The shaded region marks the 381 frames of ground truth in
each training clip of \method, whereas MiniWorld-1B was pretrained on 253-frame
clips. On Sekai the recorded camera trajectory ends at frame 1,789 and
continues with the relative motion of other held-out clips.}
\label{fig:horizon}
\end{figure}

\begin{table*}[tbp]
\centering\small
\caption{Realism over the horizon on 32 held-out rollouts per domain, in bins
of generated frames. MiniWorld-1B was pretrained on 253-frame clips and
\method{} fine-tunes on 381-frame clips. CLIP-FD is the Fr\'echet distance to real frames
in CLIP space, reported next to that of GT with the same number of frames as the
floor for each bin, and FVD uses I3D on 16-frame windows. Lower is better for both, while
brightness is best at 1 and flicker at the GT's value. On Sekai GT exists up
to frame 1,789. In a paired bootstrap over rollouts (500 resamples), \ours{}
improves on the CLIP-FD of MiniWorld-1B in every bin of both domains in
100\% of resamples, and on its FVD from frame 509 onwards on DROID and in every
bin on Sekai in at least 97\%.}
\label{tab:main}
\fontsize{9}{11}\selectfont
\setlength{\tabcolsep}{1.87pt}
\begin{tabular}{l cccccc c cccccc}
\toprule
 & \multicolumn{6}{c}{\textbf{DROID}} && \multicolumn{6}{c}{\textbf{Sekai}} \\
\cmidrule{2-7}\cmidrule{9-14}
 & 1--253 & 253--509 & 509--1k & 1k--1.8k & 1.8k--3k & 3k--4k && 1--253 & 253--509 & 509--1k & 1k--1.8k & 1.8k--3k & 3k--4k \\
\midrule
\rowcolor{RowBlue}\multicolumn{14}{l}{\textit{CLIP-FD} $\downarrow$} \\
\quad \textit{GT} & \textit{6} & \textit{10} & \textit{5} & \textit{5} & \textit{5} & \textit{5} && \textit{128} & \textit{131} & \textit{122} & \textit{110} & -- & -- \\
\quad MiniWorld-1B & 38 & 61 & 105 & 223 & 348 & 431 && 292 & 447 & 490 & 524 & 554 & 609 \\
\quad FM-only & 38 & 66 & 117 & 217 & 320 & 401 && 238 & 371 & 419 & 476 & 505 & 564 \\
\quad \ours & \best{30} & \best{42} & \best{38} & \best{39} & \best{40} & \best{44} && \best{148} & \best{181} & \best{197} & \best{213} & \best{209} & \best{226} \\
\midrule
\rowcolor{RowBlue}\multicolumn{14}{l}{\textit{FVD} $\downarrow$} \\
\quad MiniWorld-1B & 45 & 80 & 177 & 733 & 2{,}000 & 2{,}058 && 400 & 1{,}109 & 1{,}213 & 1{,}283 & 1{,}217 & 1{,}193 \\
\quad FM-only & 44 & 89 & 194 & 546 & 1{,}420 & 1{,}873 && 308 & 808 & 877 & 1{,}019 & 1{,}066 & 1{,}181 \\
\quad \ours & \best{39} & \best{66} & \best{57} & \best{62} & \best{70} & \best{81} && \best{186} & \best{249} & \best{316} & \best{358} & \best{372} & \best{441} \\
\midrule
\rowcolor{RowBlue}\multicolumn{14}{l}{\textit{Brightness (ratio to GT)}} \\
\quad MiniWorld-1B & \best{1.00} & \best{1.00} & \best{1.00} & 1.02 & 1.03 & \best{1.00} && 0.89 & 0.77 & 0.68 & 0.61 & 0.50 & 0.40 \\
\quad FM-only & 0.99 & 0.98 & 0.96 & 0.94 & 0.94 & 0.94 && 0.90 & 0.79 & 0.80 & 0.79 & 0.70 & 0.59 \\
\quad \ours & \best{1.00} & \best{1.00} & 0.99 & \best{0.99} & \best{0.98} & 0.98 && \best{1.01} & \best{1.02} & \best{1.02} & \best{1.07} & \best{1.01} & \best{0.92} \\
\midrule
\rowcolor{RowBlue}\multicolumn{14}{l}{\textit{Chunk-boundary flicker}} \\
\quad \textit{GT} & \textit{1.13} & \textit{0.98} & \textit{1.04} & \textit{1.03} & \textit{1.03} & \textit{1.04} && \textit{1.01} & \textit{1.01} & \textit{1.00} & \textit{1.02} & -- & -- \\
\quad MiniWorld-1B & 1.08 & 1.12 & 1.12 & 1.22 & 1.35 & 1.46 && 1.20 & 1.43 & 1.54 & 1.44 & 1.53 & 1.58 \\
\quad FM-only & 1.06 & \best{1.09} & \best{1.07} & 1.13 & 1.25 & 1.37 && 1.35 & 1.44 & 1.56 & 1.63 & 1.76 & 1.96 \\
\quad \ours & \best{1.09} & 1.13 & 1.10 & \best{1.11} & \best{1.11} & \best{1.11} && \best{1.03} & \best{1.03} & \best{1.04} & \best{1.09} & \best{1.10} & \best{1.13} \\
\bottomrule
\end{tabular}
\end{table*}

Figure~\ref{fig:horizon} and Table~\ref{tab:main} evaluate the three models
across 4,001 frames in both domains. Although MiniWorld-1B exhibits domain-specific
degradation patterns, its failure onset is identical: in both domains, quality
deteriorates sharply from the second bin onwards, immediately after the rollout
exceeds its 253-frame pretraining horizon, and never recovers. With \method{},
every metric on DROID remains nearly flat from the second bin through frame
4,001, and CLIP-FD on Sekai plateaus after the first 500 frames.

\paragraph{DROID.} For MiniWorld-1B, CLIP-FD rises from 38 in the first 253
frames to 431 in frames 2,997--4,001, FVD increases from 45 to 2,058 and
chunk-boundary flicker grows from 1.08 to 1.46 as saturated lattice artifacts
spread outward from the moving robot arm (Fig.~\ref{fig:teaser}). Because mean
brightness stays near 1.00, the degradation on DROID manifests as spatial
texture corruption rather than exposure drift. In contrast, \ours{} sustains
coherent rollouts beyond 4,000 frames: across all six bins, CLIP-FD remains
within 30--44 and FVD within 39--81, achieving $10\times$ lower CLIP-FD and
$25\times$ lower FVD than MiniWorld-1B in the final bin. Brightness stays within
2\% of GT and flicker stays between 1.09 and 1.13, matching the GT range of
0.98--1.13. Notably, DROID provides a stringent test of training without long
videos: its median episode is shorter than a single fine-tuning clip, yet the
evaluated rollouts extend to $16\times$ the pretraining horizon and over
$10\times$ the fine-tuning clip length.

\paragraph{Sekai.} On Sekai, MiniWorld-1B rollouts progressively darken and
dissolve into noise: relative brightness drops from 0.89 in the first 253 frames
to 0.40 in frames 2,997--4,001, CLIP-FD rises from 292 to 609, FVD exceeds 1,100
from frame 253 onwards, and flicker increases from 1.20 to 1.58. With \ours{},
CLIP-FD remains stable between 197 and 226 from frame 509 onwards and exhibits
no discontinuity at frame 1,789, where the camera transitions to concatenated
held-out trajectories. Across all bins, brightness stays between 0.92 and 1.07
of GT and flicker remains between 1.03 and 1.13. In the final bin, both CLIP-FD
(226 vs.\ 609) and FVD (441 vs.\ 1,193) are $2.7\times$ lower than those of
MiniWorld-1B. FVD still exhibits a gradual upward trend from 186 to 441 over the
4,000-frame horizon, a residual slow drift that we discuss in
\S\ref{sec:limits}. The 1,789-frame recorded-trajectory protocol under
independent noise confirms these gains in its final bin (CLIP-FD 226 vs.\ 504
and FVD 518 vs.\ 1,154 in Table~\ref{tab:ablation}).

\paragraph{Per-rollout inspection.} Manual inspection of the 96 DROID rollouts
confirms that population-level stability translates to individual trajectories:
98\% of \ours{} rollouts remain free of visible artifacts at frame 1,500, 95\%
at frame 2,500, 86\% at frame 3,500 and 79\% at frame 4,000. By contrast, in all
eight inspected rollouts of MiniWorld-1B, lattice artifacts cover a large
portion of the scene by frame 3,000. The artifacts that do appear in \ours{} are
strictly local, such as a manipulated colourful object or cloth turning into a
saturated region or a faint periodic pattern emerging on a uniform tabletop,
while the surrounding scene and robot arm remain intact. On Sekai, the
per-rollout brightness quantiles in Fig.~\ref{fig:horizon} show a similar
improvement: at frame 4,000, 88\% of \ours{} rollouts retain at least half of
their GT brightness, compared with only 19\% for MiniWorld-1B and 28\% for
FM-only.

\paragraph{Why flow matching alone does not suffice.} Although FM-only
fine-tuning uses the exact same training clips and self-denoised histories, it
barely alters long-horizon behaviour in either domain. On DROID, where FM-only
differs from \method{} solely by omitting the two distributional terms, its
final-bin CLIP-FD (401), FVD (1,873) and flicker (1.37) remain close to those of
MiniWorld-1B. On Sekai, FM-only improves initial quality but leaves the rate of
degradation unchanged: its CLIP-FD increases by 326 from the first to the last
bin, matching the 317-point rise of MiniWorld-1B and far exceeding the 78-point
increase of \ours{}, while its brightness drops to 0.59 and its flicker rises to
1.96. Training longer does not resolve this degradation: on DROID, the final-bin
CLIP-FD of FM-only is 433 at step 100 and 401 at step 225, and on the
1,789-frame Sekai protocol its final-bin CLIP-FD plateaus at 509, 479 and 490
after 100, 200 and 400 steps. Adding the distributional constraints, by
contrast, yields monotonic long-horizon improvements, reducing the final-bin FVD
on DROID from 186 at step 100 to 135, 112 and 81 at steps 150, 200 and 225.

This gap reflects three inherent limitations of the FM objective. First, FM is an
open-loop, single-step objective that regresses the velocity of one chunk at a
single noise level without evaluating how errors compound when the sampler feeds
generated outputs back into the KV cache over hundreds of steps. Second,
regressing towards the conditional mean under future uncertainty introduces a
slight bias towards a smoother, dimmer average appearance. Although negligible in
a single denoising step, this bias has a consistent sign and accumulates across
closed-loop steps into the steady darkening observed on Sekai
(Table~\ref{tab:main}). Third, even when conditioned on a self-denoised history,
FM remains anchored to clean data: the history is a single-pass re-rendering of
a real clip at most 317 frames long, and the target is its exact recorded
continuation. Consequently, FM never observes the severely drifted states visited
beyond a thousand frames, and wherever the re-rendered history disagrees with
the clean target, FM can only encourage the model to ignore the history rather
than recover from it. Context fidelity and reality matching supply this missing
closed-loop signal. Evaluated directly on free rollouts of the deployed sampler,
any drift in exposure, sharpness or scene content manifests as a statistical
discrepancy whose gradient flows into chunks generated from the model's own
degraded cache, teaching the model to actively pull drifting trajectories back
towards the real distribution. Conversely, the distributional terms cannot
replace generative anchoring: because they supervise only the final chunks of a
rollout, conditioning FM on clean history instead of self-generated history
leaves reality matching unable to arrest a gradual brightness decay (brightness
0.44 in Table~\ref{tab:ablation}).

\subsection{Action following over the horizon}
\label{sec:control}
\begin{table}[tbp]
\centering\small
\caption{Action following on the same held-out sets (higher is better).
DROID: PSNR (dB) against GT at the same frame, in the bins of
Table~\ref{tab:main}. Sekai: correlation $r$ between commanded and
VGGT-estimated yaw in frames 1,021--1,789 of the recorded trajectories
(recorded video: 0.92), and turn branches forked at frame 893 of the recorded
trajectories and at frame 3,581 of the long rollouts: the percentage that turn
in the commanded direction (dir.) and the percentage of the commanded
$70.5^\circ$ they reach (mag.). In the bootstrap of Table~\ref{tab:main}, the
DROID PSNR of \ours{} exceeds that of MiniWorld-1B from frame 253
onwards in 100\% of resamples.}
\label{tab:control}
\fontsize{9}{11}\selectfont
\setlength{\tabcolsep}{2.92pt}
\begin{tabular}{l cccccc c c *{2}{w{c}{27.9pt}} *{2}{w{c}{31.5pt}}}
\toprule
 & \multicolumn{6}{c}{DROID: PSNR to GT (dB)} && Sekai & \multicolumn{2}{c}{turn (frame 893)} & \multicolumn{2}{c}{turn (frame 3,581)} \\
\cmidrule{2-7}\cmidrule(r){9-9}\cmidrule(lr){10-11}\cmidrule(l){12-13}
 & 1--253 & 253--509 & 509--1k & 1k--1.8k & 1.8k--3k & 3k--4k && yaw $r$ & dir. & mag. & dir. & mag. \\
\midrule
MiniWorld-1B & \best{25.0} & 21.8 & 19.8 & 15.7 & 11.3 & 8.9 && 0.23 & 59 & 19 & 47 & $-$0.2 \\
FM-only & 24.1 & 21.4 & 19.6 & 16.3 & 12.9 & 10.4 && $-$0.13 & 47 & 2 & 38 & $-$2.8 \\
\rowcolor{RowBlue}\ours & 24.4 & \best{22.1} & \best{22.4} & \best{22.1} & \best{21.8} & \best{21.5} && \best{0.69} & \best{100} & \best{60} & \best{97} & \best{49} \\
\bottomrule
\end{tabular}
\end{table}
The gains in visual realism do not come at the expense of action
controllability (Table~\ref{tab:control}). \runin{DROID.} MiniWorld-1B is
accurate within its 253-frame pretraining horizon, reaching 25.0\,dB PSNR on
episodes seen during pretraining, but degrades steadily thereafter to 15.7\,dB
in frames 1,021--1,789 and 8.9\,dB in the final bin, with FM-only following a
similar trajectory (10.4\,dB). With \ours{}, PSNR remains at 22.4\,dB in frames
509--1,021 after the first forward pass through the episode, and stays at
21.5\,dB in the final bin after more than 4,000 frames of alternating
forward-backward manipulation. This long-horizon stability comes at a minor
cost of 0.6\,dB in the first 253 frames, where CLIP-FD and FVD nevertheless
improve over the base model. \runin{Sekai.} Along the recorded trajectories, the
correlation between commanded and VGGT-estimated yaw in frames 1,021--1,789
reaches 0.69 for \ours{}, compared with 0.23 for MiniWorld-1B and $-0.13$ for
FM-only, whose severely degraded frames no longer support reliable pose
estimation. When rollouts are counterfactually forked at frame 893 into left and
right turns, \ours{} turns in the commanded direction in 100\% of branches
(vs.\ 59\% for MiniWorld-1B and 47\% for FM-only) and achieves 60\% of the
commanded turn angle (vs.\ 19\% and 2\%). Crucially, camera controllability
persists through frame 4,000: in the last two horizon bins of the 4,001-frame
rollouts, \ours{} maintains yaw correlations of 0.71 and 0.75, compared with
0.06 and $-0.06$ for MiniWorld-1B and 0.05 in both bins for FM-only. When forked
at frame 3,581, 97\% of \ours{} branches still turn in the commanded direction
and reach 49\% of the commanded angle, whereas MiniWorld-1B (47\% direction,
$-0.2\%$ magnitude) and FM-only (38\% direction, $-2.8\%$ magnitude) have lost
responsiveness to the action stream.

%=============================================================================
\section{Analysis and Ablations}
\label{sec:analysis}
%=============================================================================
To understand why each component of \method{} is required, we first trace the
training dynamics of a sequence of variants that differ only in their
distributional terms, evaluating checkpoints every 25 steps with three diagnostic
probes (\S\ref{sec:marginal}--\ref{sec:feature}). To isolate the distributional
objectives, these variants share generative anchoring with the history
re-rendered to the end. We then examine how terminating the anchoring history on
real data eliminates chunk-boundary flicker (\S\ref{sec:seam}), and ablate each
constraint of the complete recipe (\S\ref{sec:ablation}). Probe metrics are
measured up to frame 933. We refer to the latent-FD variants by the subset of
rollout chunks they score: \emph{marginal} scores raw latent features without
clip centring, \emph{pooled} scores clip-centred features pooled across all $N$
rollout chunks even though gradients reach only the final $K$, and \emph{graded}
scores clip-centred features restricted to the $K$ gradient-carrying chunks so
that every scored sample receives a gradient.

\paragraph{Diagnostic probes.} The first two probes run on 250-latent
(997-frame) rollouts of the 32 held-out Sekai clips, while the third uses the
1,789-frame recorded-trajectory protocol of \S\ref{sec:setup}.
\runin{Context dependence.} For each held-out clip, we compute the mean latent
feature of its rollout over frames 765--933 and regress it onto the mean latent
feature of the clip's own ground truth over the same window across the 32 clips.
A regression slope of 1 indicates that every rollout preserves its clip-specific
appearance, whereas a slope of 0 indicates that all rollouts collapse to a
single input-independent appearance. MiniWorld-1B scores 0.92 and FM-only
fine-tuning remains within 0.77--0.92.
\runin{Gradient-statistic alignment.} At each training step, we compute the
cosine similarity between the mean-matching gradient step applied to the
gradient-carrying chunks and the true displacement vector required to move those
chunks' mean to the reference mean. A cosine near 1 indicates that the loss
drives the modifiable chunks towards the reference distribution, whereas a cosine
near 0 indicates that those chunks are being distorted to compensate for errors
in the detached prefix.
\runin{Held-out perceptual judge.} We track held-out CLIP-FD, which is never
optimised during training. When the training FD decreases while held-out CLIP-FD
rises, the model is exploiting the training feature space with degenerate
outputs rather than improving visual realism.

\begin{figure}[tbp]
\centering
\includegraphics[width=\linewidth]{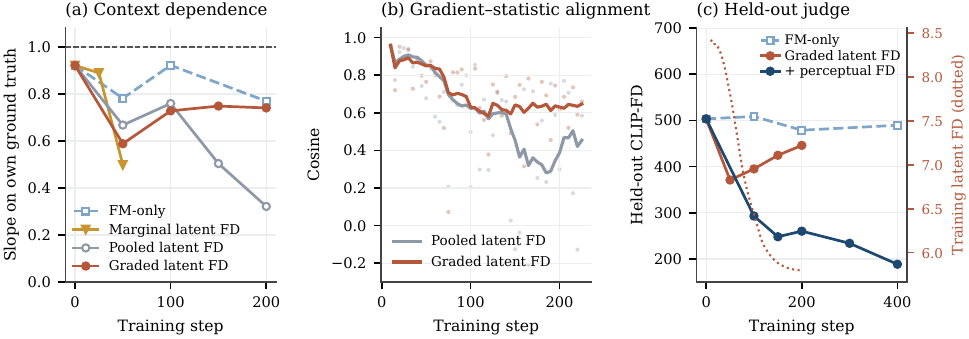}
\caption{Training dynamics of distribution matching on free rollouts. (a)
Context dependence in frames 765--933, measured by the regression slope of each
rollout's mean latent feature on that of its own ground truth across held-out
clips (1 indicates full preservation of clip-specific appearance). (b) Cosine
similarity between the mean-matching correction and the true direction needed by
the gradient-carrying samples, showing per-step training logs (dots) and a
50-step moving average (lines). (c) Held-out CLIP-FD in frames 1,021--1,789 for
FM-only, the graded latent FD alone, and the graded latent FD combined with the
perceptual FD. The dotted curve tracks the training latent FD of the graded
variant, which continues to decrease even as held-out CLIP-FD rises.}
\label{fig:dynamics}
\end{figure}

\subsection{Marginal matching trades context for a typical appearance}
\label{sec:marginal}
An unconditional marginal FD tests only whether the aggregate population of
rollout frames matches real statistics, without requiring each rollout to remain
consistent with its own initial scene. In a low-order space of per-channel
latent means and standard deviations, the model rapidly exploits this degree of
freedom, causing rollouts to decouple from their conditioning context. As shown
in Fig.~\ref{fig:dynamics}a, the context-dependence slope of the marginal
variant drops from 0.92 to 0.50 within just 50 steps, whereas FM-only stays
above 0.77 throughout training. Rather than maintaining each clip's distinct
scene across a degrading context cache, the model pulls every trajectory towards a
generic appearance that satisfies the population mean. Visually, in the
1,789-frame rollouts, scenes dissolve into a uniform reddish wash by frame 1,100
and become severely overexposed, reaching $2.1\times$ the GT brightness in
frames 1,021--1,789. Centring the rollout features on each clip's own
ground-truth mean prevents this immediate collapse, but sustaining the
improvement requires aligning the scored chunks with the truncated gradient
window.

\subsection{Gradients must reach what is measured}
\label{sec:align}
Because training rollouts span $N=40$ chunks while backpropagation is truncated
to the final $K=8$ chunks, a natural design is to pool statistics across all $N$
chunks of the rollout while differentiating only through the tail. However, doing
so creates a severe mismatch between the statistical residual and the parameter
gradient. Because the pooled mean is dominated by the $N-K$ detached prefix
chunks that the gradient cannot alter, the optimisation forces the $K$
gradient-carrying tail chunks to \emph{overshoot} in the opposite direction to
compensate for prefix drift. Consequently, by step 200, the gradient-statistic
alignment cosine of the pooled variant decays from 0.96 to 0.39
(Fig.~\ref{fig:dynamics}b) and its context-dependence slope collapses to 0.32
(Fig.~\ref{fig:dynamics}a), yielding overly smoothed far-horizon frames with
sharpness dropping to 0.30 of GT. Restricting the statistical evaluation to the
$K$ gradient-carrying chunks (the graded variant) eliminates the detached prefix
from the moment estimate so that the objective and the gradient act on the exact
same samples. At step 200, the graded variant maintains an alignment cosine of
0.63 and a context-dependence slope of 0.74. \method{} therefore decodes and
scores only the gradient-carrying tail of each rollout.

\subsection{The feature space decides what is matched}
\label{sec:feature}
Even with clip centring and gradient alignment in place, optimising the graded
latent FD alone eventually diverges from perceptual quality
(Fig.~\ref{fig:dynamics}c). While the training latent FD decreases monotonically
from 8.4 to 5.8 over 200 steps, held-out CLIP-FD in frames 1,021--1,789 reaches
a minimum of 371 at step 50 before rising to 446 by step 200. Visually, far
frames degenerate into dark, featureless fields (sharpness 0.24 of GT) or, under
the pooled variant, become covered in periodic fish-scale and diamond textures.
Because low-order latent moments do not constrain semantic structure, the
optimiser satisfies them via degenerate textures that bear no resemblance to
natural images. Introducing reality matching in the frozen DINOv2 feature space
closes this loophole. Because self-supervised ViT representations are sensitive
to high-level visual content, dark or periodically tiled frames lie far from the
real manifold, forcing the rollout to remain sharp, well-exposed and semantically
coherent. Combining the perceptual FD with the graded latent FD reduces held-out
CLIP-FD at step 200 to 215 in frames 509--1,021 (vs.\ 423 for FM-only) and to
260 in frames 1,021--1,789 (vs.\ 446 for the graded latent FD alone and 479 for
FM-only), with continued training further lowering the far-horizon CLIP-FD to
188 at step 400 (Fig.~\ref{fig:dynamics}c). Together, the two distributional
constraints ensure that extended training steadily improves long-horizon quality.

\subsection{Anchoring must end on real data}
\label{sec:seam}
\begin{table}[tbp]
\centering\small
\caption{Chunk-boundary flicker over the horizon: block-scale change across a
chunk boundary relative to the change within a latent (1: no redraw at the
chunk rate), in bins of generated frames, on 32
held-out clips. The fine-tuned rows differ only in how the history of
generative anchoring ends, and all are evaluated at 150 steps except the last.}
\label{tab:seam}
\setlength{\tabcolsep}{10.99pt}
\begin{tabular}{lccccc}
\toprule
 & 1--125 & 125--253 & 253--509 & 509--1k & 1k--1.8k \\
\midrule
\textit{GT} & \textit{1.03} & \textit{0.99} & \textit{1.01} & \textit{1.00} & \textit{1.02} \\
MiniWorld-1B & 1.16 & 1.28 & 1.38 & 1.44 & 1.45 \\
\midrule
Re-rendered to the end, $t_a=0.7$ & 1.31 & 1.28 & 1.24 & 1.32 & 1.35 \\
Re-rendered to the end, $t_a=0.4$ & 1.20 & 1.20 & 1.15 & 1.24 & 1.29 \\
\rowcolor{RowBlue}Ends on a real chunk & \best{1.07} & \best{1.03} & \best{1.03} & 1.08 & \best{1.13} \\
\rowcolor{RowBlue}\quad at 225 steps (\ours) & 1.09 & 1.05 & 1.05 & \best{1.07} & \best{1.13} \\
\bottomrule
\end{tabular}
\end{table}
While the three probes above capture distributional drift, they do not detect a
distinct temporal artifact induced by naive self-history training. When
generative anchoring re-renders the entire conditioning history up to its final
chunk, the resulting rollouts score well on frame-level realism (achieving a
CLIP-FD of 155 in frames 1--125 vs.\ 241 for MiniWorld-1B), yet exhibit prominent
periodic jumps: mid-scale structures such as tree canopies or hillsides shift
position abruptly every eight frames, coinciding with the two-latent chunk
boundary. Standard pixel-level and optical-flow temporal metrics fail to capture
this artifact because high-frequency foliage texture and local noise dominate
pixel differences. We therefore quantify chunk-boundary flicker directly as the
ratio between the mean absolute change in $16\times16$-pooled luma across a
chunk boundary and the corresponding change between adjacent frames within a
latent, averaged geometrically across clips. On natural video, this ratio lies
between 0.99 and 1.03 (Table~\ref{tab:seam}). Re-rendering the anchoring history
to the end raises flicker to 1.31 in the first 125 frames, worse than the 1.16
of pretrained MiniWorld-1B, because fine-tuning actively teaches the model to
redraw the scene at the seam between the re-rendered history and the clean FM
target (\S\ref{sec:anchoring}). This redraw is not caused by the distributional
constraints, as FM-only fine-tuning with the same re-rendered history exhibits
even higher flicker (1.36 in frames 1--125, rising to 1.71 in frames
1,021--1,789 in Table~\ref{tab:ablation}). Nor is it an inherent consequence of
self-generated history: conditioning FM on clean ground-truth history, whose
final chunk trivially matches the target continuation, keeps flicker at 1.10 in
frames 1--125. Reducing the re-noising level from $t_a=0.7$ to $t_a=0.4$
retains more of the clean signal and softens the boundary mismatch, but does not
remove it (1.20). Restoring a single clean real chunk at the end of the
self-generated history eliminates the seam entirely, reducing flicker to 1.07
(within 0.04 of GT) while preserving far-horizon realism (CLIP-FD of 268 vs.\
248 in frames 1,021--1,789 at step 150, reaching 226 at step 225 in
Table~\ref{tab:ablation}). Whereas the flicker of MiniWorld-1B worsens from 1.16
to 1.45 over the horizon as its rollouts degrade into noise, \ours{} maintains
smooth transitions between 1.05 and 1.13 across all horizon bins.

\subsection{Every constraint is necessary}
\label{sec:ablation}
\begin{table}[tbp]
\centering\small
\caption{Ablations on the 1,789-frame Sekai protocol, frames 1,021--1,789 of
the recorded trajectories. Sharpness and brightness are relative to GT (1 is
ideal), flicker is the chunk-boundary measure (GT: 1.02), and turn is the fraction
of branches forked at frame 893 that turn in the commanded direction. Rows 4--7
use the re-rendered history of row 3 and are compared with it, with rows 4--5
cumulative.}
\label{tab:ablation}
\setlength{\tabcolsep}{8.55pt}
\begin{tabular}{lccccccc}
\toprule
 & CLIP-FD & FVD & sharp. & bright. & flicker & yaw $r$ & turn \\
\midrule
\rowcolor{RowBlue}\ours{} & \best{226} & 518 & 1.33 & \best{1.03} & 1.13 & \best{0.69} & \best{1.00} \\
context fidelity in pixels & 230 & \best{336} & 0.70 & 0.79 & \best{1.03} & 0.64 & 0.97 \\
\midrule
history re-rendered to end & 248 & 463 & 1.42 & 0.94 & 1.35 & 0.57 & 0.94 \\
\quad $-$ context fidelity & 264 & 540 & 0.68 & 0.74 & 1.29 & 0.59 & 0.91 \\
\quad $-$ self history & 253 & 564 & 0.77 & 0.44 & 1.13 & 0.63 & 0.91 \\
\quad $-$ reality matching & 395 & 745 & 0.58 & 0.84 & 1.39 & 0.23 & 0.75 \\
FM-only & 479 & 1{,}002 & 0.87 & 0.73 & 1.71 & $-$0.13 & 0.47 \\
\midrule
MiniWorld-1B & 504 & 1{,}154 & 2.22 & 0.58 & 1.45 & 0.23 & 0.59 \\
\bottomrule
\end{tabular}
\end{table}
Table~\ref{tab:ablation} ablates the constraints of \method{} on the 1,789-frame
Sekai protocol, showing that removing any component reintroduces the
corresponding failure mode analysed above.

\paragraph{Reality matching.} Removing reality matching leaves only the
context-centred latent statistic to supervise the free rollout. As shown in
\S\ref{sec:feature}, the model satisfies this low-order statistic with blurry,
degenerate frames that do not resemble real images, causing CLIP-FD to rise from
248 to 395, FVD to worsen from 463 to 745, and sharpness to drop to 0.58 of
real video.

\paragraph{Context fidelity.} Dropping context fidelity leaves the optimisation
of the perceptual FD unchanged, with a DINO-FD training curve indistinguishable
from that of the full objective (1,227 vs.\ 1,222 at step 100), yet held-out FVD
worsens from 463 to 540 and brightness falls from 0.94 to 0.74. Because reality
matching is a population-level objective, it cannot detect when an individual
rollout drifts away from \emph{its own} scene as long as the overall population
remains plausible. Context fidelity provides this per-clip conditioning, and the
conditioning itself matters more than the specific space in which it is
computed. To verify this, we replace the latent statistic with 25-dimensional
low-order pixel statistics of the decoded frames of the last four chunks (RGB
means and spatial standard deviations, a $4\times4$ grid of luma means, and the
log standard deviation of the luma Laplacian at three scales), centred on each
clip's decoded ground truth in the same way against a reference of 3,600
clip-centred real segments. This pixel-space variant exhibits nearly identical
training dynamics (DINO-FD 742 vs.\ 737 at step 220), and even reduces the
latent FD that it no longer directly optimises from 7.97 to 5.83 (compared with
5.26 when optimised directly). In the far frames, it matches the latent version
on CLIP-FD (230 vs.\ 226) and improves FVD (336 vs.\ 518) and flicker (1.03 vs.\
1.13). However, its rollouts are dimmer and softer (brightness 0.79, sharpness
0.70), perform worse at intermediate horizons (CLIP-FD of 202, 212 and 221 vs.\
176, 185 and 204 across frames 125--1,021), and reach 44\% of the commanded
turn angle compared with 60\%. This trade-off reflects what low-order pixel
statistics capture: dominated by low spatial frequencies, their clip-centred
mean is easily reduced by shifting towards a dimmer, smoother appearance,
whereas the 48 VAE latent channels also constrain local structure and detail. We
therefore use the latent statistic by default and view the pixel statistic as a
practical alternative for architectures without a latent space.

\paragraph{Generative anchoring.} In generative anchoring, both the real
terminal chunk and the self-generated prefix of the history are essential.
Re-rendering the history all the way to the end produces comparably realistic
far frames (and a slightly lower FVD of 463 vs.\ 518 in the last recorded bin,
though \ours{} achieves lower FVD in every bin of the 4,001-frame rollouts, such
as 358 vs.\ 474 in frames 1,021--1,789), but it induces severe chunk-boundary
redraw, raising flicker from 1.13 to 1.35 (\S\ref{sec:seam}). Conversely,
replacing the self-generated history with clean ground truth reverts the FM term
to standard pretraining and causes rollouts to darken substantially, with
brightness falling to 0.44 compared with 0.74 under self-generated history.
Trained only on clean histories, the FM objective never exposes the model to its
own accumulated errors, and reality matching on the final chunks alone cannot
prevent a gradual decay in exposure over the rollout.

%=============================================================================
\section{Limitations}
\label{sec:limits}
%=============================================================================
\method{} constrains visual appearance rather than the magnitude of camera
motion. While turns reliably follow the commanded direction, they reach only
49\% to 60\% of the commanded angle, making an explicit pose-consistency term
on the rollout a natural complement. In addition, the perceptual feature space
operates at a coarse spatial scale, allowing some bright rollouts to develop
fine periodic textures that none of our metrics registers. Using a feature
space that is harder to satisfy at fine scales could address this issue.
Because the distributional terms operate at the population level and score only
states reached within a single 317-frame training rollout, slower drifts are
constrained only indirectly, as reflected by a minority of clips that still
fade and a gradual rise in FVD beyond 1,800 frames. Similarly, reality matching
evaluates individual frames and therefore enforces frame-level realism rather
than long-range consistency. Consequently, a scene revisited much later need
not look the same, and scenes that are rare in the training data, such as snow,
gradually drift towards the dominant content while remaining realistic. On
DROID, local artifacts such as a manipulated object turning into a saturated
blob or a periodic texture emerging on a uniform surface appear in 20\% of
rollouts by frame 4,000, mostly after frame 2,500. Because the surrounding
scene and all population metrics remain stable during these failures,
mitigating them likely requires region-level perceptual matching. Finally, the
DROID ground truth is an episode replay rather than a continuous recording, the
held-out episodes were part of the pretraining data and our experiments cover
two domains at a single model size.

\section{Conclusion}
We have presented \method{}, a method for training streaming world models for
long horizons from short clips. Sustaining long rollouts requires supervision
that remains valid across the states the model actually visits. For an
action-conditioned world model, a teacher trained on short clips cannot
reliably provide this signal, whereas the distribution of real data can.
Building on this insight, \method{} fine-tunes the deployed sampler under three
complementary constraints, where generative anchoring keeps the model a
conditional generator, context fidelity keeps each rollout faithful to its own
scene and reality matching keeps its frames realistic. Without a teacher, an
online critic or long videos, fine-tuning for at most 36 GPU-hours on 381-frame
clips turns a model that degrades after its 253 training frames into one whose
rollouts remain stable beyond 4,000 frames and continue to follow their actions
in both robot manipulation and camera-controlled exploration. Because the
training signal never requires the recorded future, the length of the training
clips no longer bounds the usable horizon of a rollout, at least up to the
4,000 frames we evaluate. Our analysis further characterises how training on
free rollouts fails through context decoupling, gradient-statistic
misalignment, feature-space exploitation and the seam between self-generated
history and real targets, which we hope will inform future work on long-horizon
world models.

\clearpage

\bibliographystyle{plainnat}
\setlength{\bibhang}{0pt}
\setlength\bibindent{0pt}
\bibliography{refs}

\end{document}